\documentclass[11pt]{article}

\usepackage[final]{acl}

\usepackage{times}
\usepackage{latexsym}
\usepackage{amsmath}
\usepackage{amssymb}

\usepackage[T1]{fontenc}
\usepackage[utf8]{inputenc}

\usepackage{microtype}
\usepackage{xspace}

\usepackage{inconsolata}

\usepackage{graphicx}
\usepackage{enumitem}
\usepackage[breakable]{tcolorbox} 
\usepackage{booktabs}
\usepackage{multirow}
\usepackage{xcolor}
\usepackage{subcaption}

\newcommand{\up}[1]{\textcolor{red!70!black}{\tiny$\uparrow$#1}}
\newcommand{\dn}[1]{\textcolor{green!45!black}{\tiny$\downarrow$#1}}

\title{
Learning Agent-Compatible Context Management for Long-Horizon Tasks
}
\author{
 \textbf{Lu Yi\textsuperscript{1*}},
 \textbf{Runlin Lei\textsuperscript{1*}},
 \textbf{Liuyi Yao\textsuperscript{2}},
 \textbf{Yuexiang Xie\textsuperscript{2}},
\\
 \textbf{Yuyang Li\textsuperscript{3}},
 \textbf{Wenhao Zhang\textsuperscript{2}},
 \textbf{Zhewei Wei\textsuperscript{1$\dagger$}},
 \textbf{Yaliang Li\textsuperscript{2$\dagger$}},
 \textbf{Jian-Yun Nie\textsuperscript{4}}
\\
\\
 \textsuperscript{1}Renmin University of China,
 \textsuperscript{2}Tongyi Lab, Alibaba Group\\
 \textsuperscript{3}Beijing University of Posts and Telecommunications,
 \textsuperscript{4}Université de Montréal
}
\def\method{AdaCoM\xspace}
\def\methods{AdaCoMs\xspace}
\def\header{\vspace{0.8mm} \noindent}
\def\bcp{BrowseComp-Plus\xspace}
\def\mcp{MCP-Bench\xspace}
\def\mcpwiki{MCP-Bench-Wiki\xspace}

\def\q4b{Qwen3-4B-Instruct\xspace}
\def\qwen{Qwen3-Max\xspace}
\def\ds{DeepSeek-V3\xspace}
\def\kimi{Kimi-K2-Instruct\xspace}
\def\gpt4omini{GPT-4o-mini\xspace}
\def\doubao{Seed-1.6-Flash\xspace}
\def\glm{GLM-4.5-Air\xspace}
\def\gemini{Gemini-2.5-Flash\xspace}
\def\oss{GPT-OSS-20B\xspace}
\def\react{ReAct\xspace}
\def\sumagent{SumAgent\xspace}
\def\memact{MemAct\xspace}
\def\sumcom{SumCoM\xspace}
\def\qwenS{Qwen\xspace}
\def\dsS{DeepSeek\xspace}
\def\kimiS{Kimi\xspace}
\def\glmS{GLM\xspace}
\def\geminiS{Gemini\xspace}
\def\doubaoS{Seed\xspace}

\begin{document}
\maketitle
\renewcommand{\thefootnote}{\fnsymbol{footnote}}
\footnotetext[1]{Lu Yi and Runlin Lei are co-first authors. Work done during internship at Tongyi Lab, Alibaba Group.}
\footnotetext[2]{Zhewei Wei and Yaliang Li are corresponding authors.}
\begin{abstract}
LLM agents increasingly face long-horizon tasks such as web search and deep research in real-world applications, where accumulated context can cause long-context degradation and reasoning failures. 
Prior work mitigates this through context management with agent-side context control or fixed strategies such as summarization, which require training the agent itself for adaptation — making it impractical for closed-source agents and ignoring that different agents may require different strategies.
We introduce Adaptive Context Management (\method), which trains an external LLM to manage the context of a frozen agent through flexible modification actions and end-to-end reinforcement learning. 
Across diverse agents on web search and deep research benchmarks, \method substantially improves performance by preserving task constraints and progress while pruning stale content. 
The learned strategies reveal a \textit{Fidelity--Reliability Trade-off}: agents with higher vanilla ReAct performance benefit from higher-fidelity context preservation, whereas lower-performing agents require more aggressive compression to stay within a reliable reasoning regime. 
Transfer experiments show that \method generalizes most effectively across agents with similar capability (measured by vanilla ReAct performance), suggesting a practical path toward reusable context managers for agent systems.
\end{abstract}

\section{Introduction}
\begin{figure}[t]
\centering
\includegraphics[width=0.5\textwidth]{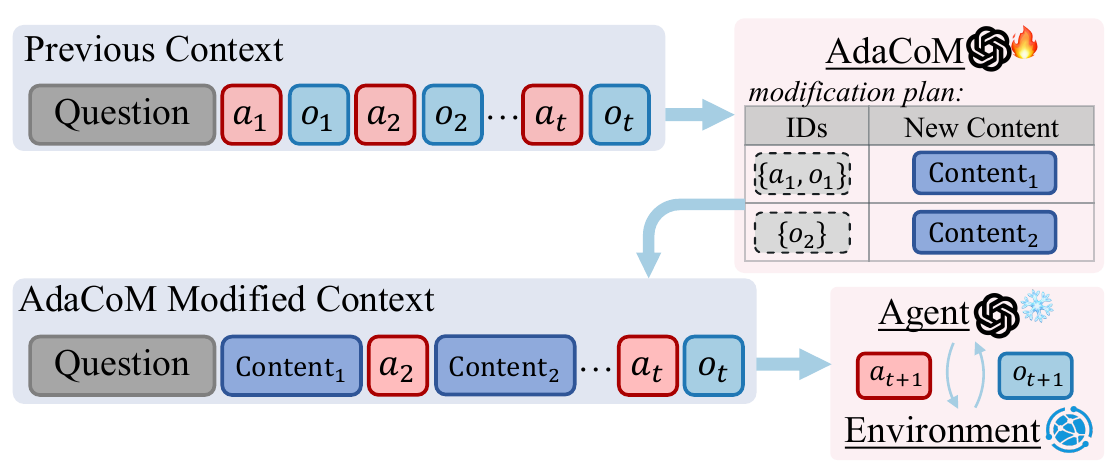}
\caption{
Overview of Adaptive Context Management (\method). Before each agent step, an external LLM manages the context presented to the frozen agent. Task feedback updates only the manager, enabling \method to discover agent-compatible context management strategies without training the underlying agent.
}
\label{fig:AdaCoM}
\vspace{-5mm}
\end{figure}

With advances in semantic understanding, tool use, and interactive decision making, general-purpose LLM agent applications such as OpenClaw and Hermes Agent have emerged~\cite{openclaw,hermes}. 
Such applications often involve long-horizon reasoning, where tasks such as answering multi-constraint search queries~\cite{wei2025browsecomp,li2025websailor} or producing deep research reports~\cite{du2025deepresearch,wang2025mcp} require many interdependent steps over a growing context. 
A central bottleneck for LLMs in such long-horizon tasks is long-context degradation.
As tool results and intermediate reasoning accumulate, stale or irrelevant content can obscure salient evidence, amplify positional bias, and make subsequent decisions less reliable~\cite{attentionsink,lossinmiddle,distraction}.

Prior work addresses this issue through \textit{context management}, but typically places the burden of managing context on the agent itself.
The agent is prompted to summarize its trajectory before acting~\cite{zhou2025mem1,chen2025iterresearch}, or to invoke context-management tools~\cite{memact}. 
Because these mechanisms require the agent to follow the new context patterns or tool-use protocols that it may not have encountered during training, prior work often couples them with agent training to realize performance gains~\cite{wu2025resum}.
This training-dependent design is poorly matched to deployment, where widely used agents are often closed-source, and training every user-selected model is infeasible. 
Moreover, predefined operations such as summarization impose a one-size-fits-all strategy despite substantial variation in agents' architectures, training data, and reasoning styles. 
This motivates a natural question: \emph{can we discover the preferred context management strategy for each agent without training the agent itself?}

We introduce {\bf Adaptive Context Management (\method)}, a framework based on two principles. 
First, \emph{architectural decoupling}: context management is handled by an external manager, typically a smaller LLM, while the agent itself remains unchanged.
This decoupling makes \method applicable even when the agent cannot be trained. 
Second, \emph{operation-level flexibility}: instead of committing to predefined operations such as summarization, the manager can freely modify any part of the context, allowing it to discover agent-compatible management strategies. 
To learn such strategies, we train the manager with reinforcement learning while keeping the agent frozen.
Figure~\ref{fig:AdaCoM} illustrates \method's process: before each agent step, the manager updates the running context, and the agent then acts on the resulting managed context.

Across diverse agents, \method substantially improves performance on web search and deep research tasks, reducing constraint forgetting, premature abandonment, and redundant exploration by retaining task-relevant context.
Further analysis of \method's learned strategies reveals a consistent \textbf{Fidelity--Reliability Trade-off}. 
Using vanilla \react~\cite{yao2022react} performance as a measure of agent capability, we find that managers for stronger agents preserve more raw trajectory context to maintain fidelity, whereas managers for weaker agents compress more aggressively to keep reasoning reliable.
This suggests that each agent has an effective context length beyond which additional raw context becomes harmful, and desirable context management must balance context fidelity against reasoning reliability. 
Further transfer experiments show that trained managers transfer most effectively between agents with comparable capability, suggesting that practitioners can reuse a trained manager across capability-similar agents without retraining the manager.

Overall, our main contributions are:
\begin{itemize}[itemsep=0.5pt,topsep=2pt]
\item We propose \method, an adaptive context management framework that decouples context management from the agent and learns agent-compatible strategies without retraining the agent itself.
\item \method substantially improves diverse agents on web search and deep research tasks, mitigating constraint forgetting, premature abandonment, and redundant exploration.
\item We identify the Fidelity--Reliability Trade-off and show that \method transfers most effectively across capability-similar agents, providing practical guidance for deploying context management in real-world agent applications.
\end{itemize}

\section{Related Work}
\header{\bf Context management for long-horizon tasks.}
Existing context management methods typically reduce long trajectories through predefined operations. Summarization methods such as IterResearch~\cite{chen2025iterresearch} and MEM1~\cite{zhou2025mem1} maintain a compact progress summary during task solving. 
Upon each observation, the agent conditions on the previous summary and the latest action-observation pair, then updates the summary and issues the next action in the same generation.
ReSum~\cite{wu2025resum} uses a separate summarization tool when the context approaches the length limit, compressing previous interaction history before the agent continues. Tool-based methods such as \memact~\cite{memact} expose context management to the agent through a pruning tool, allowing it to summarize selected historical messages, remove the originals, and append the summary. 
These methods prescribe the management operation in advance and/or require the agent to learn when and how to manage its own context, often coupling context management with agent training. 
In contrast, \method trains an external manager to manage context while keeping the underlying agent fixed.

\header{\bf Long-term memory for LLM agents.} A related but distinct line of work studies long-term memory for LLM agents~\cite{hu2025memorysurvey}. 
These methods focus on storing and retrieving information across sessions, tasks, or users, including factual knowledge and experiential memory. 
Representative systems include Mem0~\cite{chhikara2025mem0}, A-Mem~\cite{xu2026mem}, Memory-R1~\cite{yan2025memoryr1}, and G-Memory~\cite{zhang2026gmemory}. 
Our work is complementary to this direction. 
Instead of building persistent memory across conversations, we study \emph{working-memory management} within a single long-horizon task, aiming to keep agents' active context useful for task completion.

\section{Adaptive Context Management}
\label{sec:method}

This section presents \method, an adaptive context management framework for long-horizon agentic tasks. We first describe its workflow and flexible modification action space (Section~\ref{subsec:paradigm}), and then introduce the reinforcement learning procedure for training the context manager (Section~\ref{subsec:rl}).

\subsection{Flexible Context Management Paradigm}
\label{subsec:paradigm}

\header{\bf Agent's vanilla workflow.}
Given a task query $q$, a ReAct-style agent $\mathcal{A}$ maintains an accumulated context $c_t^{\mathrm{vanilla}} = (q, a_1, o_1, \ldots, a_t, o_t)$ at turn $t$, where $a_i$ contains the agent's reasoning and tool invocation and $o_i$ is the corresponding environment observation. The agent generates the next action by conditioning on the entire context: $a_{t+1} \sim \mathcal{A}(c_t^{\mathrm{vanilla}})$. This append-only workflow is common in long-horizon tasks, but suffers from long-context degradation as context accumulates.

\header{\bf \method-augmented workflow.}
\method uses an external manager to modify the agent's running context, leaving the underlying agent $\mathcal{A}$ frozen. We denote the context manager's policy by $\pi_\theta(\cdot\mid p)$, which selects structured modification actions given a management prompt $p$.

Let $\tilde{c}_{t-1}$ denote the managed context after turn $t-1$, with $\tilde{c}_0=(q)$. At turn $t$, the agent first acts on the managed context, $a_t \sim \mathcal{A}(\tilde{c}_{t-1})$, and the environment returns an observation $o_t$. We append the new action and observation to obtain the pre-management context $c_t = \mathrm{Append}(\tilde{c}_{t-1}, a_t, o_t)$. The manager samples a structured modification action $m_t \sim \pi_\theta(\cdot \mid p_t)$, with $p_t=\mathcal{P}(c_t)$, where $\mathcal{P}$ constructs the prompt with the management instruction and output schema. Applying $m_t$ to $c_t$ produces the next managed context $\tilde{c}_t$. The next agent action is then generated from $\tilde{c}_t$.

\header{\bf \method's action space.}
A central design goal of \method is operation-level flexibility. Rather than committing to a predefined operation such as summarization, we formulate context management as general-purpose modification over a message sequence. At each turn, the pre-management context $c_t$ is represented as an ordered list of messages with unique message IDs.

The manager action $m_t$ is a structured list of modification operations, expressed in JSON as
\[
m_t = \bigl[ \delta_t^{(1)}, \delta_t^{(2)}, \ldots, \delta_t^{(n_t)} \bigr],
\]
where each operation $\delta_t^{(j)}$ selects one or more messages and specifies how they should be rewritten, deleted, or merged; it can also specify the role of the resulting message.
Each operation contains four fields: 
(1) $\mathrm{ids}^{(j)}$, the IDs of the targeted messages; 
(2) $\mathrm{role}^{(j)} \in \{\textsc{system}, \textsc{user}, \textsc{assistant}\}$, the role of the resulting message; 
(3) $\mathrm{justification}^{(j)}$, a short rationale that elicits the manager's reasoning about the modification, encouraging higher-quality edits; it is removed before the managed context is shown to the agent; and
(4) $\mathrm{new\_content}^{(j)}$, the resulting content, where empty content deletes the targeted messages and non-empty content rewrites or merges them. 
Messages not targeted by any operation are copied unchanged, and an empty action list leaves the context unchanged.
The full manager prompt is provided in Appendix~\ref{app:prompts}.

This action space supports diverse context-management operations beyond a fixed strategy. \method can remove stale information, condense verbose evidence, merge related messages, or leave the context unchanged when fidelity is important. 

\subsection{Training the Context Manager}
\label{subsec:rl}

\header{\bf Problem formulation.}
We formulate manager learning as a Markov decision process (MDP) induced by the frozen agent and the environment. At each manager step $t$, the current pre-management context $c_t$ is formatted into a prompt $p_t=\mathcal{P}(c_t)$. The manager policy emits a structured modification action $m_t \sim \pi_\theta(\cdot \mid p_t)$, which is then applied to $c_t$ to update the managed context. A rollout induces a sequence of manager decision points $\tau=((p_1,m_1),\ldots,(p_T,m_T))$. The rollout terminates when the frozen agent emits a final answer or exceeds the maximum number of interaction steps. The trajectory-level outcome reward is the task reward of the agent's final answer.

\header{\bf Training overview.}
We first use supervised fine-tuning (SFT) to initialize the manager with the required output format, and then optimize it with Group Relative Policy Optimization (GRPO)~\citep{grpo}. For each query, we sample multiple rollouts from the current manager policy and evaluate the final answer produced by the frozen agent. For tasks with deterministic answers, an LLM judge compares the final answer against the ground truth and produces a binary score; for open-ended tasks, it scores the answer according to a task-specific rubric.

\header{\bf Process reward design.}
We additionally introduce process rewards to improve credit assignment in long-horizon interactions. We use rule-based process rewards computed directly from trajectories, without invoking an additional LLM judge. If the managed context exceeds the given context length limit, we apply a \emph{token penalty} to the manager step that produced the over-length context. 
When the agent issues two consecutive tool calls with the same tool name and parameters, we apply a \emph{redundant-action penalty} to the intervening manager action, treating repetition as a proxy for insufficient preservation of useful information.
We also introduce \emph{format penalties} for structurally invalid manager outputs, including JSON parsing failures, nonexistent message IDs, and missing fields. Finally, when a benchmark provides task-specific intermediate signals, we incorporate them as positive process rewards, as detailed in Section~\ref{sec:exp}.

\header{\bf Two-level advantage estimation.}
In our setting, outcome rewards and process rewards are defined at different granularities and may have different numerical scales, so directly adding their raw values can make optimization sensitive to reward scaling.
Motivated by prior work on reward normalization in RL-based LLM training~\citep{ding2025empowering, liu2026gdpo}, we first convert them into separately normalized advantages before combining them to improve training stability.
Specifically, for each query $q$, we sample a group of $G$ rollouts $\{\tau_i\}_{i=1}^{G}$ from the old manager policy $\pi_{\theta_{\mathrm{old}}}$. To index rollouts within the group, we add an index $i$ to the notation. Thus, $p_{i,t}$ and $m_{i,t}$ denote the management prompt and manager action at step $t$ of rollout $i$, respectively. The $i$-th rollout is $\tau_i=((p_{i,1},m_{i,1}),\ldots,(p_{i,T_i},m_{i,T_i}))$, where $T_i$ is the number of manager steps in rollout $i$, and each action satisfies $m_{i,t}\sim\pi_{\theta_{\mathrm{old}}}(\cdot\mid p_{i,t})$. Let $R_i$ denote the terminal outcome reward of rollout $\tau_i$, and let $Q_{i,t}$ denote the aggregate process reward assigned to manager step $t$ in rollout $i$, obtained by summing all applicable process-reward components described above. 

We compute advantages at two granularities.
The \emph{task-level} advantage assigns the same normalized outcome reward to each step in rollout $\tau_i$:
$$
A^R_i = \frac{R_i - \mu_R}{\sigma_R + \varepsilon},
$$
where $\mu_R$ and $\sigma_R$ are the mean and standard deviation of $\{R_j\}_{j=1}^{G}$. The \emph{step-level} advantage normalizes process rewards over all steps in the group:
$$
A^Q_{i,t} = \frac{Q_{i,t} - \mu_Q}{\sigma_Q + \varepsilon},
$$
where $\mu_Q$ and $\sigma_Q$ are computed over $\{Q_{i,t}: i=1,\ldots,G,\; t=1,\ldots,T_i\}$.

We combine the two levels as $A_{i,t}=A^R_i+\alpha A^Q_{i,t}$, where $\alpha$ controls the balance between outcome and process supervision. We then re-normalize $A_{i,t}$ over all steps in the group:
$
\hat A_{i,t}
=
({A_{i,t}-\mu_A})/({\sigma_A+\varepsilon}),
$
where $\mu_A$ and $\sigma_A$ are computed over $\{A_{i,t}: i=1,\ldots,G,\; t=1,\ldots,T_i\}$. When $\sigma_R=0$, all rollouts in the group receive the same outcome reward. The task-level term then vanishes, and the step-level term provides the only learning signal within such same-outcome groups.

\header{\bf Policy optimization.}
We optimize $\pi_\theta$ with a PPO-style clipped surrogate over manager-emitted tokens. Let $u$ index the tokens in $m_{i,t}$, and let $r_{i,t,u}(\theta)$ be the token-level importance ratio between $\pi_\theta$ and $\pi_{\theta_{\mathrm{old}}}$. The objective is
$$
\mathcal{J}(\theta)
=
\mathbb{E}
\left[
\frac{1}{Z}
\sum_{i,t,u}
\min\!\left(
r_{i,t,u}(\theta)\hat A_{i,t},
\bar r_{i,t,u}(\theta)\hat A_{i,t}
\right)
\right],
$$
where $\bar r_{i,t,u}(\theta)=\mathrm{clip}(r_{i,t,u}(\theta),1-\epsilon,1+\epsilon)$, the expectation is over $q\sim\mathcal{D}$ and rollouts sampled from $\pi_{\theta_{\mathrm{old}}}$, and $Z=\sum_{i=1}^{G}\sum_{t=1}^{T_i}|m_{i,t}|$ is the total number of manager-emitted tokens.

\section{Experiments and Analysis}\label{sec:exp}
\subsection{Experimental Setup}
\begin{table*}[t]
\centering
\resizebox{\textwidth}{!}{
\begin{tabular}{@{}l cccccc@{}}
\toprule
\textbf{Agent / Setting}
& \textbf{\react}
& \textbf{\sumagent}
& \textbf{\memact}
& \textbf{\sumcom}
& \textbf{\method w/o train.}
& \textbf{\method} \\
\midrule
CM Model     & -- & agent itself & agent itself & Qwen3-4B-Inst. & Qwen3-4B-Inst. & Qwen3-4B-Inst. \\
CM Trained   & -- & $\times$ & $\times$ & $\checkmark$ & $\times$ & $\checkmark$ \\
\midrule
\qwen
& 27.78
& 26.67\,\dn{4.0\%}
& \textbf{37.33}\,\up{34.4\%}
& 32.22\,\up{16.0\%}
& 21.11\,\dn{24.0\%}
& \underline{36.67}\,\up{32.0\%} \\

\kimi
& 18.56
& 30.44\,\up{64.0\%}
& 16.89\,\dn{9.0\%}
& \underline{33.78}\,\up{82.0\%}
& 28.02\,\up{51.0\%}
& \textbf{36.20}\,\up{95.0\%} \\

\glm
& \underline{32.56}
& 11.56\,\dn{64.5\%}
& 32.00\,\dn{1.7\%}
& 26.44\,\dn{18.8\%}
& 21.33\,\dn{34.5\%}
& \textbf{35.33}\,\up{8.5\%} \\

\ds
& 17.78
& 19.56\,\up{10.0\%}
& 16.67\,\dn{6.2\%}
& \underline{25.11}\,\up{41.2\%}
& 17.57\,\dn{1.2\%}
& \textbf{26.19}\,\up{47.3\%} \\
\midrule
\textbf{Avg.}
& 24.17
& 22.06\,\dn{8.7\%}
& 25.72\,\up{6.4\%}
& \underline{29.39}\,\up{21.6\%}
& 22.01\,\dn{8.9\%}
& \textbf{33.60}\,\up{39.0\%} \\
\bottomrule
\end{tabular}
}%
\caption{%
  \textbf{Mean@3} (\%) on BrowseComp-Plus across four agents.
  Each non-\react cell reports the absolute score and the relative change against the same-agent \emph{\react} baseline.
  Bold and underline indicate the best and second-best results in each row.
  The average is computed over agents.
  Pass@3 results are reported in Appendix~\ref{app:exp_res}.
}
\label{tbl:trained_bcp}
\vspace{-3mm}
\end{table*}

\header{\bf Benchmarks.}
We evaluate on two representative long-horizon tasks: web search and deep research. 
For web search, we use \bcp~\citep{browsecompplus}, a BrowseComp-derived benchmark~\citep{wei2025browsecomp} with a verified corpus for controlled evaluation. The agent is provided with two tools: \texttt{search(query, top\_k)}, which retrieves the top-$k$ relevant documents, and \texttt{get\_document(doc\_id)}, which returns the corresponding document content. We create a disjoint training/test split of 680/150 instances and use Qwen3-Embed-8B as the retriever, following~\citet{foldagent}. The maximum number of iterations is set to 35, and GPT-4o (2024-11-20) is used as a judge to produce a binary correctness score.

For deep research, we follow the data construction procedure of \mcp~\citep{wang2025mcp} and build a Wikipedia-MCP-based benchmark, denoted as \mcpwiki. The agent can use nine Wikipedia MCP tools, covering search and content retrieval. Following the original benchmark protocol, we use Claude Opus 4.6 as the judge. While the original benchmark evaluates reports along six dimensions, we focus on three dimensions most relevant to context management: task fulfillment (TF), information grounding (IG), and parallelism and efficiency (PE).
The overall score is computed as $0.4\,\mathrm{TF} + 0.4\,\mathrm{IG} + 0.2\,\mathrm{PE}$. Scores on all evaluation dimensions are provided in Appendix~\ref{app:exp_res}.

For \bcp, we report \textbf{mean@3}, the average accuracy over three independent runs, and \textbf{pass@3}, whether at least one run answers correctly. For \mcpwiki, we report \textbf{mean@3}, the average rubric-based score over three runs.

\header{\bf Implementation.}
\method is initialized from \q4b, briefly warmed up with SFT to learn the edit format, and then trained with the RL procedure in Section~\ref{subsec:rl}. For the SFT warm-up, we use GPT-5 (2025-08-07) and Claude Opus 4.6 to generate edit trajectories for \bcp and \mcpwiki, respectively, with \qwen serving as the frozen agent on the corresponding training set. For RL training, we build on Trinity-RFT~\citep{pan2025trinity} with group size $G=8$ for both benchmarks. 
We set the manager's own input context-window length to 32,768 tokens and its maximum output length to 4,096 tokens.
For \bcp, we additionally use positive process rewards as the benchmark provides gold and key document IDs. When a search result contains a gold or key document, we assign a \emph{gold-doc-found} or \emph{key-doc-found} reward to the preceding manager step, crediting the context management that led the agent to issue the successful search. Additional experimental settings and process reward parameters are deferred to Appendix~\ref{app:exp}.

\header{\bf Baselines.}
We compare \method against five baselines.
(i)~\emph{\react}: the vanilla \react loop without context management.
(ii)~\emph{\sumagent} (Summarization Agent): a summarization paradigm used in MEM1~\citep{zhou2025mem1} and IterResearch~\citep{chen2025iterresearch}, where the agent conditions on the original question, previous summary, and latest tool interaction to update the summary and issue the next action.
(iii)~\emph{\memact}~\citep{memact}: the agent can invoke a prune tool to summarize selected historical messages, remove them, and append the generated summary.
(iv)~\emph{\sumcom} (Summarization Context Manager): a trained external \q4b manager summarizes the context after each agent step while the agent remains frozen. This ablates our flexible management actions by replacing them with a fixed summarization operation; we adapt prompts from ReSum~\citep{wu2025resum} and use the same training procedure as \method.
(v)~\emph{\method w/o training}: the original \q4b backbone used as the context manager with the same modification action space but without SFT or RL.

\paragraph{Agent naming.}
For readability, we use shortened names for frozen agents when unambiguous: Qwen for \qwen, DeepSeek for \ds, Kimi for \kimi, GLM for \glm, Gemini for \gemini, and Seed for \doubao.

\begin{figure*}[t]
\centering
\begin{minipage}[b]{0.40\textwidth}
\centering
\includegraphics[width=\linewidth]{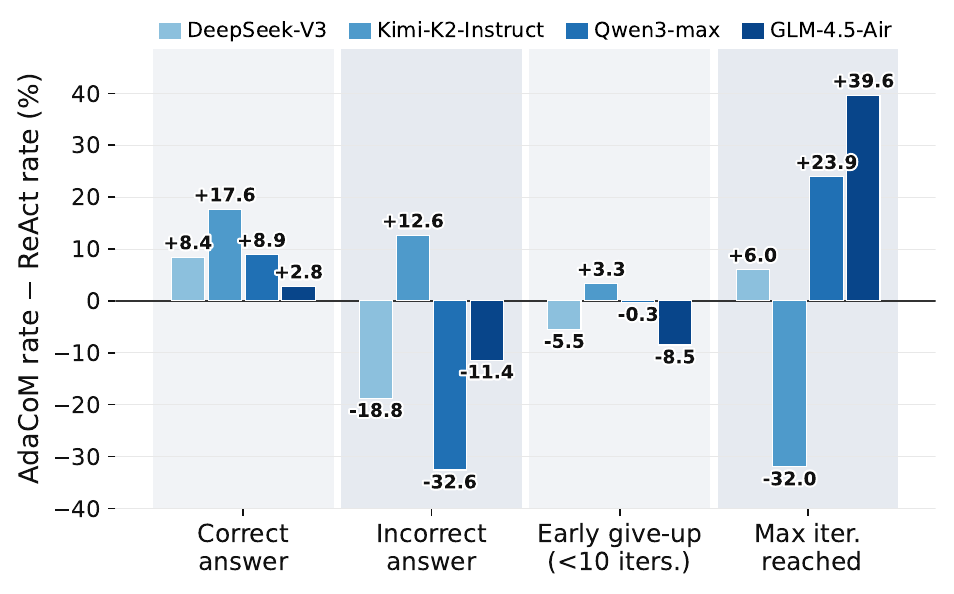}
\captionof{figure}{Per-agent trajectory outcome distribution shifts (\%) on \bcp, computed as \method minus \react.}
\label{fig:failure_modes}
\end{minipage}\hfill
\begin{minipage}[b]{0.58\textwidth}
\centering
\stepcounter{figure}\setcounter{subfigure}{0}
\begin{subfigure}[b]{0.49\linewidth}
\centering
\includegraphics[width=\linewidth]{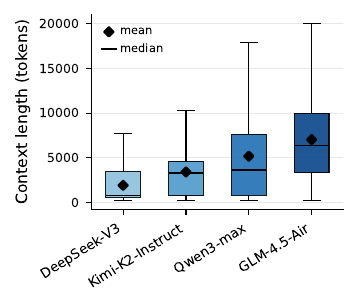}
\vspace{-7mm}
\phantomsubcaption\label{fig:cm_diff_ctx_hist}
\end{subfigure}\hfill
\begin{subfigure}[b]{0.49\linewidth}
\centering
\includegraphics[width=\linewidth]{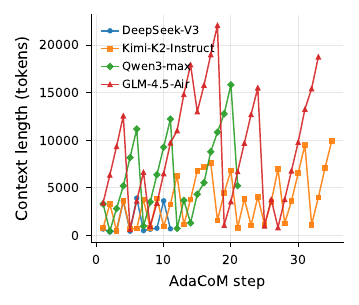}
\phantomsubcaption\label{fig:cm_diff_task_curves}
\vspace{-7mm}
\end{subfigure}
\addtocounter{figure}{-1}
\captionof{figure}{Post-\method agent context length across four agents, each paired with its self-trained \method: \textbf{(a)} per-modification distribution; \textbf{(b)} per-step trace on task 199.}
\label{fig:cm_diff}
\end{minipage}
\vspace{-2mm}
\end{figure*}

\subsection{Main Results}

Table~\ref{tbl:trained_bcp} reports the results on \bcp. 
For \method, we train a separate context manager for each target agent, including \qwenS, \dsS, \kimiS, and \glmS. 
Overall, \method consistently improves over the vanilla \react baseline across all four agents, achieving an average relative gain of 39.0\% and a 95.0\% relative gain on \kimiS. 
These results demonstrate the effectiveness of learning an external context manager while keeping the underlying agent unchanged.

We further compare against several alternatives to understand where the gains come from.
First, \method w/o training performs inconsistently and is worse than \react on average, indicating that the manager must be trained to induce effective context-management strategies.
Second, \sumcom improves some agents but degrades \glmS, suggesting that a fixed summarization operation is not sufficiently compatible with varied agent backbones.
Third, self-management baselines such as \memact and \sumagent also exhibit unstable behavior and do not outperform \react across all agents.
This supports the hypothesis that asking the agent to manage its own context may require agent-side training, since the resulting summaries may not be reliably usable by the original agent~\citep{wu2025resum}.
Notably, \memact marginally outperforms \method on \qwenS. We attribute this to \qwenS's strong intrinsic agentic capability, which enables it to invoke the pruning tool effectively even without additional training. However, this advantage does not generalize to other agents, whereas \method provides consistent gains across agents.

Table~\ref{tbl:trained_mcp} reports the results on \mcpwiki, which serves as a supplementary evaluation beyond web search. 
\mcpwiki tests \method in deep research tasks requiring multi-step MCP tool invocation, evidence gathering, and long-form report synthesis. 
Due to the higher cost of deep research rollouts and evaluation, we evaluate two representative agents, \kimiS and \dsS. 
The results show \method improves both agents over \react, achieving 9.0\% and 22.3\% relative gains on \kimiS and \dsS, respectively, indicating that the advantages of learned external context management can be extended to tool-use-intensive tasks like deep research.

\begin{table}[t]
\centering
\resizebox{\columnwidth}{!}{
\begin{tabular}{@{}l ccc@{}}
\toprule
\textbf{Agent / Setting}
& \textbf{\react}
& \textbf{\method w/o train.}
& \textbf{\method} \\
\midrule
\kimi
& 55.05
& 44.35\,\dn{19.4\%}
& \textbf{60.01}\,\up{9.0\%} \\

\ds
& 47.51
& 47.82\,\up{0.7\%}
& \textbf{58.09}\,\up{22.3\%} \\
\midrule
\textbf{Avg.}
& 51.28
& 46.09\,\dn{10.1\%}
& \textbf{59.05}\,\up{15.2\%} \\
\bottomrule
\end{tabular}
}%
\caption{%
  \textbf{Mean@3} on \mcpwiki across two agents.
  Each non-\react cell reports the absolute score and the relative change against the \emph{\react} baseline.
}
\label{tbl:trained_mcp}
\vspace{-3mm}
\end{table}

\subsection{Analysis: How \method Helps Agents}
\label{sec:analysis_how_cm_helps}

\header{\bf Trajectory outcome taxonomy.}
To understand how learned context management helps agents solve long-horizon tasks, we analyze how \method changes outcome distributions on the \bcp test set.
We categorize trajectories into four interpretable outcomes:
(i)~\textit{Correct answer}, where the agent terminates with a correct final answer;
(ii)~\textit{Incorrect answer}, where the agent terminates with a definitive but wrong final answer;
(iii)~\textit{Early give-up}, where the agent stops within ten iterations without providing a definitive answer; and
(iv)~\textit{Max iter. reached}, where the agent reaches the iteration limit without a valid answer. We omit non-definitive stops after more than ten iterations to keep the analysis focused.

\header{\bf Outcome shifts and underlying causes.}
Figure~\ref{fig:failure_modes} reports the change in outcome ratios from \react to \method, computed as the ratio under \method minus the ratio under \react.
Overall, \method increases the proportion of correct answers across all backbones compared to \react, while also shifting the composition of failure modes.
For \qwenS, \dsS, and \glmS, \method reduces \textit{Incorrect answer} and \textit{Early give-up}, suggesting fewer wrong commitments and early give-ups.
Paired trajectory inspection shows that these reductions are mainly tied to mitigating \textit{constraint forgetting} and \textit{premature abandonment}. 
In failed \react trajectories, agents often commit to candidates satisfying only part of the task requirements or give up after several ineffective searches despite remaining solvable directions.
In successful \method trajectories, the manager often maintains a compact state message immediately after the user task message, listing task requirements, unresolved constraints, useful evidence, current leads, rejected candidates, and ineffective queries.
These elements help the agent verify candidates against all requirements and keep searching rather than committing to wrong answers or abandoning the task.

For \kimiS, \method sharply reduces \textit{Max iter. reached}, indicating fewer long, unproductive trajectories.
This reduction is mainly tied to mitigating \textit{redundant exploration}: in failed \react trajectories, \kimiS often repeatedly issues identical or near-identical tool calls.
On the \bcp test set, an average of 42.6\% of \kimiS's tool-use steps per task are repetitive.
\method reduces such loops by recording prior attempts and pruning redundant or unhelpful tool results from the active context, thereby lowering the reasoning burden and making repeated exploration less likely.

Importantly, we do not explicitly prompt \method on what to record or how to organize the context.
\method learns to record and update task requirements, search progress, and prior attempts as needed, helping agents maintain an adaptive task state rather than merely shortening the context.

\begin{figure*}[t]
\centering
\vspace{-3mm}
\includegraphics[width=0.95\textwidth]{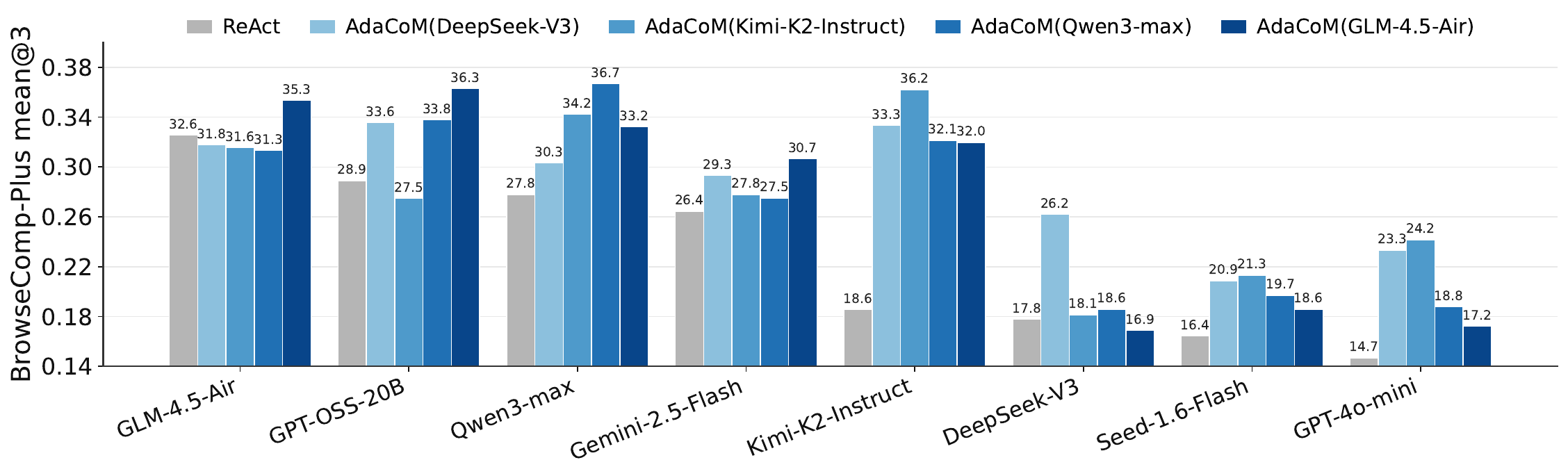}
\vspace{-3mm}
\caption{Cross-agent transfer of \method on \bcp. Each group on the $x$-axis is a target agent; bars within a group show \bcp \textbf{mean@3} under \react and various trained \methods.}
\label{fig:transfer}
\vspace{-3mm}
\end{figure*}

\subsection{Analysis: Specific Strategies across Agents}
\label{sec:cm_diff}

We further compare the learned context management behavior across agents on the \bcp test set.
Figure~\ref{fig:cm_diff} shows that the learned strategies differ systematically across agents.
The mean post-management context length forms a clear trend: about $1.9$K tokens for \dsS, $3.4$K for \kimiS, $5.2$K for \qwenS, and $7.0$K for \glmS.
Notably, this trend aligns with the agents' vanilla \react performance on \bcp, ordered from lower to higher as \dsS, \kimiS, \qwenS, and \glmS.
In other words, managers for stronger \react agents preserve more context, while managers for weaker agents compress more aggressively.
Per-task context-length curves further reveal two characteristic strategies.
For \glmS and \qwenS, \method follows a \emph{tiered management} strategy: it lets the context grow by retaining raw tool results and performs occasional batched compression.
For \dsS and \kimiS, \method follows an \emph{eager distillation} strategy: it compresses nearly every round to keep the context short, with \kimiS preserving slightly more raw context than \dsS.

\header{\bf Fidelity--Reliability Trade-off.}
These empirical strategies reveal a {Fidelity--Reliability Trade-off}.
Using vanilla \react performance as a proxy for agent capability, stronger agents such as \glmS and \qwenS can exploit longer raw contexts, so \method intervenes later and more lightly; weaker agents such as \dsS and \kimiS benefit from shorter, distilled contexts, so \method compresses more frequently.
This reflects a balance between fidelity and reliability: preserving raw trajectories retains fine-grained evidence, but beyond an agent's effective context length, additional raw content can harm reasoning.
Conversely, aggressive compression improves reliability but may discard details.
Thus, effective context management is not a fixed operation such as summarization, but an agent-compatible strategy that preserves as much fidelity as possible while keeping reasoning reliable.

\header{\bf Relation to fixed summarization.}
This also explains the behavior of \sumcom in Table~\ref{tbl:trained_bcp}.
Since \sumcom always applies a fixed summarization pattern, it resembles the eager distillation strategy and brings larger gains for weaker \react agents such as \dsS and \kimiS.
However, it is less suitable for agents such as \glmS and \qwenS that can benefit from preserving more raw context, which explains why \sumcom lags behind \method and even degrades \glmS.

We provide a detailed case study showing how \method maintains task constraints and search progress while adopting agent-specific compression strategies in Appendix~\ref{app:case_study}.

\section{Cross-Agent Transferability}
\label{sec:transfer}

In practical agent applications, users may choose various agents for tasks of varying difficulty, making it costly to train a separate context manager for every agent. We therefore investigate whether a trained \method can be reused across agents without retraining on each target agent.

\header{\bf Experimental setup and overall results.}
On \bcp, we evaluate the four trained managers from Section~\ref{sec:exp}, each trained with one source agent among \dsS, \kimiS, \qwenS, and \glmS, on eight target agents. The target set includes the four source agents and four held-out agents that never participate in manager training: \oss, \geminiS, \doubaoS, and \gpt4omini.
Figure~\ref{fig:transfer} reports the accuracy of each target agent under \react and \react equipped with each of the four trained managers. Trained managers improve over \react in most configurations: 23 of 28 cross-agent pairs, plus all 4 self-trained pairs, achieve positive gains, with an average relative improvement of 25.0\% across all 32 pairs and 22.1\% across the 28 cross-agent pairs alone. The largest cross-agent gain reaches 79.6\% on \kimiS under the \dsS-trained manager, comparable to its 95.0\% self-trained gain and showing that non-self managers can approach self-trained performance. These results show that trained context managers are not narrowly tied to their source agents and can transfer broadly to unseen target agents.

We also observe similar cross-agent transfer improvements on \mcpwiki, with detailed results deferred to Appendix~\ref{app:exp_res}, suggesting that transfer is not limited to web search but also extends to tool-use-intensive deep research tasks.

\header{\bf Capability-based transfer patterns.}
We next ask what factors predict successful transfer on \bcp. A clear pattern is that effective transfer is closely related to source-target capability proximity, measured by \react baseline performance. Higher-baseline target agents, including \glmS, \qwenS, \geminiS, and \oss, generally benefit more from managers trained on higher-baseline sources such as \glmS or \qwenS, whereas lower-baseline targets such as \kimiS, \dsS, \doubaoS, and \gpt4omini tend to prefer managers trained on lower-baseline sources such as \dsS or \kimiS. This pattern is consistent with the Fidelity--Reliability Trade-off in Section~\ref{sec:cm_diff}: agents with higher baseline performance can benefit from managers that preserve more raw context, while agents with lower performance tend to benefit from more aggressive compression and distillation.

\header{\bf Exceptions and agent-specific factors.}
Capability proximity is a strong but incomplete predictor of transfer. In several exceptional cases, managers trained on agents with similar capability do not yield the best transfer performance. For example, \dsS gains little from \method~(\kimiS): although their \react baselines are similar, \dsS prefers concise working memories with key findings and document IDs, whereas \method~(\kimiS) tends to record detailed search histories. Similarly, \glmS gains little from managers trained on other agents because these managers compress too aggressively. For \geminiS, \method~(\qwenS) underperforms \method~(\kimiS) because \geminiS does not use \method~(\qwenS)'s report-style working memory effectively. Overall, capability proximity is a useful first-order rule, while compatibility between the manager's modification style and the target agent's reasoning style further affects transfer quality. Detailed trajectory analyses are provided in Appendix~\ref{app:transfer_exception}.

\header{\bf Practical implications.}
These results suggest a practical deployment strategy for context management with closed-source agents. Instead of training a separate manager for every agent, practitioners can train a small set of managers for representative agents and reuse them across capability-similar targets. Capability proximity provides a useful first-order heuristic, while agent-specific style compatibility remains important when selecting the best manager within a capability group.

\section{Conclusion}

We introduced \method, an adaptive context management framework that trains an external manager to manage an agent's running context while keeping the underlying agent frozen. With a flexible modification action space and RL training, \method learns agent-compatible context management strategies. Experiments show that \method consistently improves diverse agents and can transfer to unseen agents with similar baseline capability. Our analysis further reveals a Fidelity--Reliability Trade-off: effective context management should preserve as much task-relevant information as possible while keeping the agent within a context regime where its reasoning remains reliable.

% \clearpage
\section*{Limitations}

\paragraph{Evaluation scope.}
Our experiments cover two long-horizon task families, web search (\bcp) and deep research (\mcpwiki), both centered on knowledge-intensive search. We do not evaluate \method on other long-horizon settings such as code agents, embodied agents, or longform writing, where the structure of useful context may differ. In addition, due to the higher cost of deep-research rollouts and evaluation, \mcpwiki uses only two agents, so conclusions on that benchmark are based on a small agent pool.

\paragraph{Manager capacity.}
Our manager is initialized from a relatively small \q4b model. This makes \method practical and inexpensive, but a 4B manager may have limited capacity to losslessly preserve or compress evidence for very strong agents whose reasoning relies on high-fidelity context. We therefore do not extensively evaluate newer and stronger closed-source agents, as such a capacity mismatch could underestimate the framework's potential. For example, the relatively small gains on \glmS may partially reflect this mismatch in addition to its already-high baseline. Due to computational constraints, we do not train larger context managers in this work. For stronger target agents, future work can pair them with more capable managers, or draw on the strategies discovered by \method, such as tiered management for high-capability agents and eager distillation for lower-capability agents, to inform the design of context-management capabilities in future agent training.

\paragraph{Inference overhead.}
\method introduces an additional manager inference at every agent step, increasing per-rollout token cost and wall-clock latency compared with vanilla \react. Our goal in this work is to discover agent-compatible management patterns rather than to optimize runtime efficiency. Notably, our findings on stronger agents suggest a natural mitigation: \method learns a tiered management strategy that intervenes only occasionally (Section~\ref{sec:cm_diff}), indicating that in deployment the manager need not be invoked at every step. Triggering it only every few rounds or when the context approaches a length threshold could substantially reduce the overhead while preserving the benefits.

\paragraph{KV cache efficiency.}
\method modifies the agent's running context between steps, which can reduce the effectiveness of KV cache reuse~\cite{pagedattention}. This limitation is shared by many context-management paradigms, including summarization-based methods, because changing earlier context tokens invalidates cached states. In this work, we focus on discovering agent-compatible context-management strategies and therefore do not optimize for KV cache preservation. However, our findings suggest a possible direction for cache-aware context-management: for stronger agents that can reliably process longer contexts, the manager can preserve raw context for more steps and compress only near the agent's effective context length. Such tiered management may reduce the frequency of cache-breaking operations while still preventing long-context degradation.

\bibliography{custom}

\appendix\label{sec:appendix}

\begin{table*}[t]
\centering
\resizebox{\textwidth}{!}{%
\setlength{\tabcolsep}{4pt}
\begin{tabular}{@{}l cccccc@{}}
\toprule
 & \textbf{\react} & \textbf{\sumagent} & \textbf{\memact} & \textbf{\sumcom} & \textbf{\method w/o train.} & \textbf{\method} \\
\midrule
CM Model     & --    & agent itself    & agent itself & Qwen3-4B-Inst. & Qwen3-4B-Inst. & Qwen3-4B-Inst. \\
CM Trained   & --    & $\times$        & $\times$     & $\checkmark$   & $\times$       & $\checkmark$ \\
\midrule
\qwen          & 38.00 & 38.67\,\up{1.8\%}  & \underline{50.67}\,\up{33.3\%} & 42.67\,\up{12.3\%}             & 30.67\,\dn{19.3\%} & \textbf{52.00}\,\up{36.8\%} \\
\kimi   & 29.33 & 44.67\,\up{52.3\%} & 30.00\,\up{2.3\%}              & \underline{45.33}\,\up{54.6\%} & 42.75\,\up{45.8\%} & \textbf{53.02}\,\up{80.8\%} \\
\glm        & 44.67 & 20.00\,\dn{55.2\%} & \underline{46.00}\,\up{3.0\%}  & 36.00\,\dn{19.4\%}             & 32.67\,\dn{26.9\%} & \textbf{50.67}\,\up{13.4\%} \\
\ds        & 26.67 & 28.67\,\up{7.5\%}  & 22.67\,\dn{15.0\%}             & \underline{37.33}\,\up{40.0\%} & 27.70\,\up{3.9\%}  & \textbf{39.97}\,\up{49.9\%} \\
\midrule
\textbf{Avg.}      & 34.67 & 33.00\,\dn{4.8\%}  & 37.33\,\up{7.7\%}              & \underline{40.33}\,\up{16.3\%} & 33.45\,\dn{3.5\%}  & \textbf{48.92}\,\up{41.1\%} \\
\bottomrule
\end{tabular}}%
\caption{%
  \textbf{Pass@3} (\%) on BrowseComp-Plus across four agents.
  Each non-\react cell shows the absolute score and the relative change versus
  the same-agent \emph{\react} baseline ($\uparrow$ = improvement, $\downarrow$ = decline).
}
\label{tbl:trained_bcp_pass3}
\end{table*}

\section{Additional Experimental Settings}\label{app:exp}
\subsection{Available Tools}

\paragraph{BrowseComp-Plus.}
The \bcp\ agent has access to two tools:
\begin{itemize}[leftmargin=*]
  \item \texttt{search(query)}: queries the corpus and returns a ranked list of document snippets, each with a unique document ID and a relevance score.
  \item \texttt{get\_document(doc\_id)}: fetches the full-text content of a document by its ID.
\end{itemize}

\paragraph{MCPWiki.}
The \mcpwiki\ agent connects to a Wikipedia MCP server.
Each tool name is prefixed with the server identifier (e.g., \texttt{Wikipedia\_\_search\_wikipedia}).
The nine available tools are:
\begin{itemize}[itemsep=0.5pt,topsep=2pt]
  \item \texttt{search\_wikipedia(query)}: keyword search returning a list of matching article titles and snippets.
  \item \texttt{get\_article(title)}: full-text retrieval of a Wikipedia article.
  \item \texttt{get\_summary(title)}: retrieves a concise summary of an article.
  \item \texttt{summarize\_article\_for\_query(title, query)}: returns a query-focused summary of an article.
  \item \texttt{summarize\_article\_section(title, section)}: summarizes a specific section of an article.
  \item \texttt{extract\_key\_facts(title)}: extracts key facts from an article, optionally focused on a topic.
  \item \texttt{get\_related\_topics(title)}: returns topics related to an article via its links and categories.
  \item \texttt{get\_sections(title)}: lists the section structure of an article.
  \item \texttt{get\_links(title)}: retrieves all links contained within an article.
\end{itemize}

\paragraph{Shared \texttt{finish} tool.}
Both benchmarks share a \texttt{finish(answer, explanation)} tool that the agent calls to terminate the episode and submit its final answer.

On \mcpwiki, whenever \texttt{get\_article} results are appended to the context, \method applies an \texttt{extract} operation before the agent's next step. This operation keeps task-relevant sentences and discards irrelevant content, preventing unbounded context growth from long Wikipedia pages. To ensure that the manager learns this behavior, we include manager invocations involving \texttt{extract} in both the SFT warm-up data and RL training trajectories. The extraction prompt is shown in Table~\ref{tab:prompt_extract}.

\subsection{Detailed Parameter Settings}

\paragraph{Process rewards.}
We use the same rule-based process reward parameters across agents and benchmarks unless otherwise specified. The \emph{token penalty} is set to $-0.8$, the \emph{redundant-action penalty} to $-0.4$, and the \emph{format penalty} to $-0.5$. For \bcp, we additionally use positive document-retrieval rewards: \emph{gold-doc-found} is set to $0.6$, and \emph{key-doc-found} is set to $0.3$. Positive and negative process rewards are accumulated when multiple rules are triggered at the same step.

\paragraph{Hyperparameters.}
We set the process reward weight $\alpha$ to $0.1$, the rollout batch size to $32$ trajectories, and the training batch size to $768$ manager-step samples. The learning rate is $5\times 10^{-6}$, the KL coefficient is $0.006$, and the PPO clipping range is $0.2$. We do not use an entropy bonus. During rollouts, the manager sampling temperature is set to $1.0$; we found that lower temperatures more easily lead to degenerate repetitive outputs. The maximum number of rollout iterations is $35$. We train for at most $40$ epochs and apply early stopping if the evaluation score does not improve for $20$ rollout steps.
We train the SFT warm-up for five epochs on \bcp and two epochs on \mcpwiki.

\paragraph{Context length. }
Under \react, each agent uses its own default maximum context length as provided by the model API: $256$K for \qwen, $128$K for \glm, $256$K for \kimi, and $128$K for \ds. Under \method, since the context manager is configured with a $32$K-token context budget, the agent's effective input is also capped at $32$K tokens regardless of the underlying model's native context window.

\section{Construction of the \mcpwiki Dataset}
\label{app:mcpwiki_data}

In the \mcpwiki task, we follow the original benchmark in creating tasks with a more fine-grained filtering process~\cite{wang2025mcp}. 
Construction proceeds in two stages: an \emph{upstream} stage that creates a clean pool of tasks, and a \emph{downstream} stage that turns those tasks into the three datasets used for training and evaluation. 
The final data contain $4{,}740$ SFT samples, $1{,}000$ RL training tasks, and a held-out test set of $150$ tasks, as summarized in Table~\ref{tab:app_mcp_data}.

\subsection{Upstream: task synthesis pool}
\label{sec:dataset:upstream}

The upstream pipeline produces, validates, and de-duplicates a pool of Wikipedia tasks. 
Each task is a long, structured natural-language request that requires multiple coordinated calls to a Wikipedia MCP server. 
Every task carries three fields used downstream: \texttt{task\_id}, \texttt{task\_description}, and \texttt{dependency\_analysis}.

The pipeline has four steps:

\begin{itemize}[leftmargin=*]
  \item \textbf{Multi-model generation.} Tasks are drafted by five frontier LLMs in a round-robin schedule, including Claude Opus 4.6, Gemini 3.1 Pro, GLM-5, MiniMax M2.7, and GPT-5 (2025-08-07).
    Each task-generation prompt requires the model to reason about cross-tool dependencies, embed concrete arguments, and avoid topics already covered by an updating blacklist for diversity.
    The blacklist is refreshed every $100$ accepted tasks: the top-$5$ most frequently quoted entities from the accepted pool so far are appended, which steers subsequent generations away from saturated topics and encourages topical diversity across rounds.
  \item \textbf{Cross-model evaluation.} Three judges, drawn from the same model pool but excluding the generator, score each candidate on a 1--10 \emph{solvability} scale. 
    Only tasks judged solvable (score $\ge$ 7) by at least two judges survive.
  \item \textbf{Dry-run validation.} A separate executor agent
    (Claude Opus 4.6) attempts the task end-to-end against the live Wikipedia MCP server. 
    A task is considered successful only if every tool call returns without error, the number of calls is not too low, and the final answer exceeds 100 characters to ensure it is non-trivial.
  \item \textbf{Log analysis and filtering.} A GLM-5 judge reads the full execution log of each passing run and assigns an accept/reject decision plus a 1--10 quality score covering completion, well-definedness, tool usage, and answer quality.
    Tasks that fail dry-run validation or are rejected by the log judge are discarded.
\end{itemize}

For the held-out evaluation pool, the same pipeline is re-run with one addition: the blacklist is \emph{initialized} with frequently mentioned entities mined from training-side task descriptions, so that test topics avoid training topics from the first generation round.
The dynamic refresh mechanism then continues during held-out generation.
The result is passed through a rule-based pre-filter that drops tasks shorter than 800 characters, near-duplicates (string similarity $\geq 0.7$), and topics appearing in more than five tasks.
After the four-step pipeline plus cross-round de-duplication and error removal, the upstream training pool retains $2{,}517$ tasks out of roughly $3{,}500$ raw tasks. 
The held-out evaluation pool retains $879$ candidate test tasks out of another $1{,}000$ raw generations.

\subsection{Downstream: from pool to datasets}
\label{sec:dataset:downstream}

The downstream stage runs the upstream pool through a Wikipedia agent with the context manager enabled, and turns the resulting trajectories into the deployed SFT, RL, and test datasets.

\paragraph{Benchmark rollouts.}
Every task in the $2{,}517$-task training pool is executed once with the context manager enabled. 
Claude Opus 4.6 serves as the context manager, and Qwen3-Max serves as the task-solving agent. 
The per-step context cap is $24{,}576$ tokens, with an extract threshold of $2{,}000$ tokens. 
Each rollout produces a research answer together with the full sequence of context-manager invocations, which becomes the raw material for SFT extraction.

\paragraph{Task-level filter.}
A first pass rejects rollouts that are clearly broken or uninformative.
We remove rollouts whose completion status is not \texttt{completed}, whose answers are shorter than 50 characters or begin with \texttt{"Error"}, or whose task-fulfillment score is below 2. 
The survivors are de-duplicated by text embeddings (cosine $\geq 0.92$), and the remaining tasks are vetted by a Claude Opus 4.6 judge that scores completeness, naturalness, sufficiency, and grounding on a 1--5 scale.
A task is kept only if all four scores exceed 2 and their mean is at least 3.5.

\paragraph{SFT samples.}
For each task that passes the filter, every context-manager invocation in its rollout is converted into a candidate \texttt{(prompt, response)} sample. 
Each sample is labeled by a coarse action category: \texttt{modify}, \texttt{delete}, \texttt{no\_change}, or \texttt{extract}. 
These labels are used only for organizing SFT data and do not define a separate action space. 
The \texttt{modify} label covers non-empty context rewriting operations; \texttt{delete} corresponds to setting \texttt{new\_content} to empty for the targeted messages; \texttt{no\_change} corresponds to outputting an empty modification list; and \texttt{extract} corresponds to the long-tool-result extraction operation described in Appendix~\ref{app:exp}, where an oversized tool result is first extracted before being inserted into the context.

The samples are also screened to ensure syntactic validity: responses must be parseable JSON, shorter than 12K characters, and, for \texttt{modify} actions, reduce the targeted content by at least 20\%.
Near-duplicates are removed by a MinHash sketch of the input text. 
A final ratio-control step anchors the dataset on informative \texttt{modify} samples and caps the other three action types relative to the \texttt{modify} count: \texttt{delete} and \texttt{no\_change} are each limited to at most $10\%$ of the \texttt{modify} count, and \texttt{extract} is limited to at most $30\%$.
This yields a final composition of roughly $73\%$ \texttt{modify}, $20\%$ \texttt{extract}, and $7\%$ \texttt{delete}/\texttt{no\_change} combined, for a total of $4{,}740$ samples used as input to the SFT phase.

\paragraph{RL tasks.}
RL training is sampled from the same upstream training pool as SFT but operates at the \emph{task} level rather than the per-invocation level.
Starting from the same rollouts used for SFT extraction, we keep only tasks that completed with task-fulfillment $\geq 3$, ran between 10 and 39 reasoning rounds to exclude trivially short and overly long traces, and whose descriptions are neither prefix near-duplicates of an already-kept task nor written in an overly prescriptive style.
From the remaining pool we sample $1{,}000$ tasks, retaining only the fields needed at training time: \texttt{task\_id}, \texttt{task\_description}, and \texttt{dependency\_analysis}.

\paragraph{Test tasks.}
Evaluation tasks are drawn from the held-out upstream test pool rather than the training pool, eliminating task-source overlap. 
We additionally require the upstream log-analysis quality score to be at least 6, the executed trajectory to contain between 30 and 150 tool calls, and the reference answer to exceed 500 characters. 
We also remove prefix near-duplicates and overly prescriptive descriptions. 
A random sample of $150$ tasks satisfying these criteria forms the test set used in our experiments.

\subsection{Summary}
\label{sec:dataset:summary}

Table~\ref{tab:app_mcp_data} shows the funnel from raw task generation to deployed datasets.
The training-side and test-side pools are produced by independent runs of the upstream pipeline with non-overlapping topic blacklists; together with cross-round de-duplication, this ensures that the test split has no task-level overlap with SFT or RL.

\begin{table}
\centering
\caption{Data in the \mcpwiki experiment.}
\label{tab:app_mcp_data}
\resizebox{\linewidth}{!}{
\begin{tabular}{lcc}
\toprule
\textbf{Stage} & \textbf{Train pool} & \textbf{Test pool} \\
\midrule
Raw multi-model generations & $\sim$$3{,}500$ & $1{,}000$ \\
After upstream filter \& dedup & $2{,}517$ & $879$ \\
\midrule
\multicolumn{3}{l}{\textbf{Final deployed datasets}} \\
SFT samples  & $4{,}740$ & --- \\
RL tasks & $1{,}000$ & --- \\
Test tasks & ---     & $150$ \\
\bottomrule
\end{tabular}
}
\end{table}

\section{Additional Experiment Results and Analysis}\label{app:exp_res}
\subsection{Additional Experiment Results}
\paragraph{Per-agent training pass@3 results on BrowseComp-Plus.}
Table~\ref{tbl:trained_bcp_pass3} reports the pass@3 score of \method and baselines on \bcp across four agents. The trends are consistent with the mean@3 results in Table~\ref{tbl:trained_bcp}: \method achieves the best average pass@3 and improves all four agents over \react. In contrast, \method w/o training and self-management baselines remain unstable across agents, while \sumcom improves several agents but still lags behind \method on average. These results further support the benefit of training a flexible external context manager.

\begin{figure}[t]
\centering
\includegraphics[width=\linewidth]{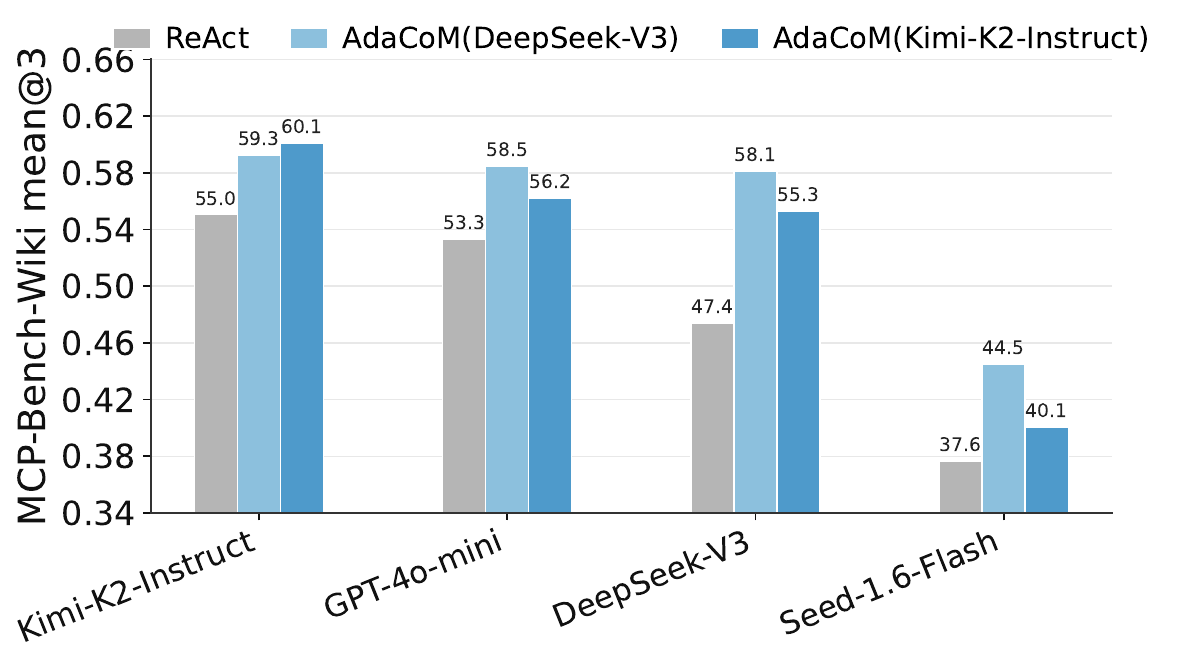}
\vspace{-6mm}
\caption{Cross-agent transfer of \method on \mcpwiki. Each group on the $x$-axis is a target agent; bars within a group show \mcpwiki \textbf{mean@3} under \react and the two trained \methods.}
\label{fig:transfer_mcp}
\vspace{-3mm}
\end{figure}

\paragraph{Cross-agent results on \mcpwiki.}
Figure~\ref{fig:transfer_mcp} shows the transfer effect on \mcpwiki. Specifically, we test the two managers trained with \ds and \kimi on four target agents: \ds, \kimi, \gpt4omini, and \doubao. The results show that both trained managers improve every target agent over \react. The consistent gains across both source managers and all target agents provide additional evidence that the transfer behavior observed on \bcp is not benchmark-specific.

\paragraph{Per-dimension scores of \mcpwiki.}
The original \mcp judge scores each rollout along six rubric dimensions: task fulfillment (TF), information grounding (IG), tool appropriateness (TA), parameter accuracy (PA), dependency awareness (DA), and parallelism and efficiency (PE). In our setting, we use the three dimensions most related to context management: TF, IG, and PE. The overall score used for both training and evaluation is computed as $0.4\times\mathrm{TF}+0.4\times\mathrm{IG}+0.2\times\mathrm{PE}$. Table~\ref{tbl:transfer_mcp_dims} reports the per-dimension breakdown.

For \gpt4omini, \ds, and \doubao, trained managers improve all three dimensions over \react, indicating that context management helps not only final task completion and grounding, but also the efficiency of tool use. \kimi shows a different pattern: \method improves the higher-weighted TF and IG dimensions while sacrificing PE relative to \react. This suggests that the manager helps \kimi gather and ground more useful information, but may encourage more tool use or longer investigation, reducing parallelism and efficiency. As TF and IG receive larger weights in the overall score, this trade-off still leads to a higher final score.

\subsection{Ablation Study on Process Reward}
We ablate the effect of process rewards in \method. Specifically, we train the managers for \ds and \qwen with the process-reward weight set to $\alpha=0$, so that the advantage is computed only from the outcome reward. Table~\ref{tbl:ab} compares the original \method with this variant on \bcp. Removing process rewards consistently reduces performance for both agents. For \ds, mean@3 drops from $26.19\%$ to $24.38\%$; for \qwen, it drops from $36.67\%$ to $33.33\%$. These results indicate that process rewards provide useful intermediate supervision beyond the sparse final-answer reward.

\begin{table}[h]
\centering
\setlength{\tabcolsep}{6pt}
\begin{tabular}{@{}lcc@{}}
\toprule
\textbf{Setting} & \textbf{\ds} & \textbf{\qwen} \\
\midrule
\method w/o pr. & 24.38 & 33.33 \\
\method                    & \textbf{26.19} & \textbf{36.67} \\
\bottomrule
\end{tabular}
\caption{%
  \textbf{Mean@3} (\%) of the original \method and \method without process reward on \bcp.
}
\label{tbl:ab}
\end{table}

% \begin{figure}[t]
% \centering
% \includegraphics[width=\linewidth]{plots/capability_compensation.pdf}
% \caption{Best-source \method behaviour vs target-agent capability (no-memory mean@3) on \bcp. Each point is one agent paired with its best transferred \method. \textbf{Top}: average agent prompt input length (tokens). \textbf{Bottom}: average number of raw \texttt{tool\_result} messages preserved in the agent context.}
% \label{fig:capability_compensation}
% \end{figure}

\begin{table*}[t]
\centering
\resizebox{0.65\textwidth}{!}{%
\setlength{\tabcolsep}{6pt}
\begin{tabular}{@{}ll cccc@{}}
\toprule
\textbf{Agent} & \textbf{Setting} & \textbf{TF} & \textbf{IG} & \textbf{PE} & \textbf{Overall} \\

\midrule
\multirow{4}{*}{\kimi}
  & \react                & 52.5 & 51.0 & \textbf{68.2} & 55.0 \\
  & \method w/o train.   & 40.7 & 50.1 & 40.0 & 44.4 \\
  & \method~(\ds)        & 56.6 & 60.4 & 62.4 & 59.3 \\
  & \method~(\kimi)      & \textbf{58.9} & \textbf{61.6} & 59.0 & \textbf{60.0} \\
\midrule
\multirow{4}{*}{\gpt4omini}
  & \react                & 45.0 & 58.2 & 60.3 & 53.3 \\
  & \method w/o train.   & 45.3 & 61.0 & 35.8 & 49.7 \\
  & \method~(\ds)        & \textbf{51.0} & \textbf{63.8} & \textbf{62.5} & \textbf{58.5} \\
  & \method~(\kimi)      & 48.0 & 63.4 & 58.3 & 56.2 \\
\midrule
\multirow{4}{*}{\ds}
  & \react                & 42.1 & 48.0 & 57.3 & 47.5 \\
  & \method w/o train.   & 41.3 & 58.6 & 39.2 & 47.8 \\
  & \method~(\ds)        & \textbf{49.1} & \textbf{64.4} & \textbf{63.5} & \textbf{58.1} \\
  & \method~(\kimi)      & 45.8 & 62.2 & 60.3 & 55.3 \\
\midrule
\multirow{4}{*}{\doubao}
  & \react                & 23.1 & 50.1 & 41.8 & 37.7 \\
  & \method w/o train.   & 26.0 & 48.8 & 34.9 & 36.9 \\
  & \method~(\ds)        & \textbf{31.6} & \textbf{55.2} & \textbf{48.8} & \textbf{44.5} \\
  & \method~(\kimi)      & 26.6 & 52.0 & 43.3 & 40.1 \\
\bottomrule
\end{tabular}
}%
\caption{
Per-dimension mean@3 (\%) of \method and baselines on \mcpwiki. 
We report task fulfillment (TF), information grounding (IG), and parallelism and efficiency (PE), with \textbf{Overall} computed as $0.4\times\mathrm{TF}+0.4\times\mathrm{IG}+0.2\times\mathrm{PE}$. 
Bold indicates the best result within each target agent and column. 
}

\label{tbl:transfer_mcp_dims}
\end{table*}

\subsection{Analysis: Transfer Exceptions and Agent Specific Factors}
\label{app:transfer_exception}

We inspect exceptional transfer cases on \bcp to better understand factors beyond baseline capability. Although capability proximity explains the dominant transfer pattern, some managers trained on capability similar agents do not yield the best performance because their modification style is not well matched to the reasoning style of the target agent.

\paragraph{\dsS and \kimiS prefer different working memory styles.}
Although \dsS and \kimiS have similar \react baselines, \dsS gains little from \method~(\kimiS). Trajectory inspection shows that \method~(\dsS) and \method~(\kimiS) organize the working memory differently. \method~(\dsS) tends to preserve concise key findings and document IDs that directly support the current search direction. In contrast, \method~(\kimiS) records a more detailed search history, including issued queries, returned document IDs, and summaries of document contents. This suggests that \dsS is more compatible with compact and exact clues, whereas \kimiS benefits more from detailed attempt histories. As a result, a manager trained for \kimiS can introduce unnecessary context for \dsS even though the two agents have similar baseline capability.

\paragraph{\glmS is sensitive to aggressive compression.}
\glmS also gains relatively little from managers trained on other agents. These managers do convert some \react failures into successes, but they also introduce regressions on tasks that \react already solves. We find that the main issue is excessive compression: managers trained on other agents often summarize or delete evidence more aggressively than \glmS prefers. Since the \q4b manager may not paraphrase evidence losslessly due to the limited capability, such compression can remove fine-grained information that \glmS could otherwise exploit from raw context. In contrast, \method~(\glmS) compresses only occasionally and in larger batches, better preserving the information density required by \glmS.

\paragraph{\geminiS is less compatible with report style memory.}
For \geminiS, \method~(\qwenS) underperforms \method~(\kimiS) despite \qwenS and \geminiS having similar \react baseline performance. We find that \method~(\qwenS) often produces working memory in a structured report style, containing previous queries and summarized results. When \geminiS reads this report, it tends to mirror the report format in its own response instead of continuing the search, which can cause the agent to give up early without a valid answer. This case shows that transfer quality can depend on whether the manager's modification style is compatible with the target agent's reasoning behavior, beyond capability proximity alone.

\section{Case Study}\label{app:case_study}

% ===== Previous (concise) version, kept for reference =====
We trace how \glm and \ds solve task 996 (a multi-constraint identity-resolution question) to illustrate how \method helps the agents complete the task and the distinct context-management strategies underlying their behavior.
Both agents reach the correct answer, but via very different paths.
\glm's \method intervenes 11 times, and in 10 of those 11 turns it modifies only a single $\sim$1\,K-token ``working note'' message (the index-1 message, placed right after the user question), while every raw tool result is left untouched. In the first three rounds, since the tool results do not contain useful clues, the working note holds a clear list of constraints about the target identity and indicates that ``no suitable candidate identified yet''. When related documents are retrieved in the fourth round, \method adds ``Current lead'' (containing the discovered clues and corresponding document IDs) and ``Pending verification'' (the constraints not yet satisfied) into the working note. In subsequent rounds, \method continues to collect new clues into the working note when they surface, or leaves the context unchanged when none arise. In the final round, in addition to keeping the working note, it performs a batched deletion that removes useless tool results and merges several stale tool results into a summary. \glm then successfully returns the answer along with verifications for all constraints.

In contrast, \ds's \method intervenes 18 times and in nearly every round either replaces the latest tool result with a paraphrase or deletes it outright. Concretely, \method maintains the refined constraints and current progress in the index-1 message throughout all rounds. Compared with the working note for \glm, this note contains more details, including the executed search queries and returned document IDs. For the latest tool result, \method often removes it in the same round, before the agent sees it in the next step. Raw results are retained only when they contain evidence for unresolved constraints, and even then for at most three rounds before being paraphrased or deleted.

\section{Code Availability}\label{app:code}

The code and data for this work are publicly available at \url{https://anonymous.4open.science/r/AdaCoM-8864/}.

\section{Artifact Licenses and Terms of Use}\label{app:licenses}

This work builds on several open-source artifacts, each used solely for non-commercial academic research.
\textbf{Trinity-RFT}~\citep{pan2025trinity}, the reinforcement learning training framework used to train the context manager, is released under the Apache 2.0 License.
\textbf{AgentScope}~\citep{gao2025agentscope}, used to implement the overall agent workflow, is likewise released under the Apache 2.0 License.
\textbf{MEM1}~\citep{zhou2025mem1}, from which we derive the \sumagent baseline, is released under the MIT License.
\textbf{MCP-Bench}~\citep{wang2025mcp}, used to construct the \mcpwiki training and evaluation set, and \textbf{BrowseComp-Plus}~\citep{browsecompplus}, used for both training and evaluation, are publicly available on GitHub; neither repository specifies an explicit open-source license at the time of use.
All artifacts are used in accordance with their intended research purposes.

Regarding data privacy, BrowseComp-Plus and MCP-Bench are constructed from publicly available web sources and benchmark corpora.
We did not collect any new human-subject data.
We manually inspected a random sample of training and evaluation instances and found no content that names or uniquely identifies private individuals, nor any offensive material.

\section{AI Assistant Usage Disclosure}\label{app:llm}

In the preparation of this manuscript, AI writing assistants (Claude, ChatGPT) were used to assist with proofreading and language polishing.
All scientific claims, experimental designs, analyses, and conclusions are the sole responsibility of the authors.

\section{Adopted Prompts}\label{app:prompts}

This section lists all prompts used in our system.
Table~\ref{tab:prompt_agent} gives the agent system prompts for \bcp\ and \mcpwiki.
Table~\ref{tab:prompt_cm} gives the \method\ context manager prompt; the placeholder \texttt{\{\{Background\}\}} is filled with a short benchmark-specific paragraph (Table~\ref{tab:prompt_background}) that describes the agent's tools and the information-retention priority for that benchmark — this is the only part of the \method\ prompt that differs between \bcp\ and \mcpwiki.
Tables~\ref{tab:prompt_judge_bcp} and~\ref{tab:prompt_judge_mcp} give the judge prompts used to score agent outputs on \bcp\ and \mcpwiki, respectively; both follow the original benchmark judge designs~\citep{browsecompplus,wang2025mcp}, with the \mcpwiki\ prompt augmented with finer-grained percentage-based scoring rubrics to improve score discrimination.
Table~\ref{tab:prompt_extract} gives the extraction prompt applied on \mcpwiki\ when \texttt{get\_article} or \texttt{get\_sections} results are appended to the context.

\begin{table*}[ht]
\caption{Agent system prompts for \bcp\ and \mcpwiki.}
\label{tab:prompt_agent}
\begin{tcolorbox}[colback=blue!5, colframe=black]

\textbf{\bcp\ Agent System Prompt}

\smallskip
You are a deep research agent. You need to answer the given question by interacting with a search engine, using the \texttt{search} and \texttt{get\_document} tools provided. Please perform reasoning and use the tool step by step, in an interleaved manner.

\smallskip NOTE:
\begin{itemize}[nosep, leftmargin=1.5em]
  \item You should always call one tool at a time. Use short keywords as the query for the \texttt{search} tool call.
  \item You should first provide your reasoning process (your chain of thought) before each tool call step.
  \item When you have a definitive answer or cannot progress further, call the \texttt{finish} tool to provide your final answer.
\end{itemize}

\bigskip
\textbf{\mcpwiki\ Agent System Prompt}

\smallskip
You are an AI assistant that can use various tools to complete tasks. You have access to tools from multiple MCP servers. Each tool is prefixed with its server name (e.g., \texttt{Wikipedia\_\_search\_wikipedia}).

\smallskip NOTE:
\begin{itemize}[nosep, leftmargin=1.5em]
  \item You should always call tools to find the information you need to complete the task, but no more than five tools at a time.
  \item When you have completed the task or cannot progress further, call the \texttt{finish} tool to provide your final answer.
\end{itemize}

\end{tcolorbox}
\end{table*}

\begin{table*}[ht]
\caption{Background information injected at the \texttt{\{\{Background\}\}} placeholder in the \method\ prompt (Table~\ref{tab:prompt_cm}).}
\label{tab:prompt_background}
\begin{tcolorbox}[colback=blue!5, colframe=black]

\textbf{\bcp\ Background Information}

\smallskip
\textbf{Task Background Information}

Agent Mission: The agent is tasked with answering complex questions by iteratively searching a database for essential clues and evidence.

Operational Workflow \& Tools:
\texttt{search} takes a query and returns a list of document snippets.
\texttt{get\_document} takes a specific document ID and retrieves the full-text content.

Core Requirement (Data Traceability): Each document is identified by a unique ID. To ensure the agent can provide verifiable citations and maintain information provenance, it is critical to preserve the document IDs associated with important information.

\bigskip
\textbf{\mcpwiki\ Background Information}

\smallskip
\textbf{Task Background}

The agent is responsible for executing complex tasks by leveraging tools across multiple MCP servers.

\textbf{Information Retention Strategy}: Retain information at a level of detail appropriate to the task at hand.
If the task requires generating structured output (e.g., reports, summaries), preserve comprehensive details.
Otherwise, retain only key findings and critical clues necessary to complete the task.

\end{tcolorbox}
\end{table*}

\begin{table*}[ht]
\caption{\method\ context manager prompt. \texttt{\{\{Background\}\}} is filled with the benchmark-specific background information in Table~\ref{tab:prompt_background}; \texttt{\{\{full\_memory\}\}} and \texttt{\{\{token\_usage\_ratio\}\}} are filled at runtime.}
\label{tab:prompt_cm}
\begin{tcolorbox}[colback=blue!5, colframe=black]

You are a \textbf{Memory Modifier} specialized in optimizing agent context windows by compressing, summarizing, or removing content while preserving task-critical information.

\smallskip\textbf{Core Objective}

Analyze the agent's context and generate specific memory modifications that:
(1) maintain only the necessary information for the current task;
(2) safely compress or remove unnecessary content;
(3) preserve logical continuity and dependencies.

\smallskip\textbf{Input}

\textbf{1. Agent Context}: The complete context window including the original task description, previous agent actions and results, and any existing summaries or hints from prior rounds.

\textbf{2. Token Usage Ratio}: Current token usage (0\% to 100\% scale).

\texttt{\{\{Background\}\}}

\smallskip\textbf{Memory Modification Process}

Follow this reasoning chain:
(1) \textbf{Analyze Context}: Understand the task progress and the agent's goal in the current round.
(2) \textbf{Apply Strategy}: Apply the compression strategy to the specific context.

\smallskip\textbf{Output Format} \quad Return \texttt{\{"modifications":[]\}} if no changes are needed.

\begin{verbatim}
{
  "modifications": [
    {
      "ids":           ["<msg_id_1>", "<msg_id_2>", ...],
      "role":          "user" | "assistant",
      "justification": "<reason for this modification>",
      "new_content":   "<replacement text, or empty string \"\" to remove>"
    }
  ]
}
\end{verbatim}

\smallskip\textbf{Field Specifications}
\begin{itemize}[nosep, leftmargin=1.5em]
  \item \textbf{ids}: One or more \emph{consecutive} message IDs to modify together.
  \item \textbf{role}: Perspective for the new content. Use \texttt{"user"} to increase agent attention to critical information.
  \item \textbf{justification}: Your justification for the modification.
  \item \textbf{new\_content}: Replacement content or summary; empty string \texttt{""} for complete removal.
\end{itemize}

\smallskip\textbf{Strategy Guidelines}

Your role is to manage the agent's context window, \emph{not} to guide its problem-solving. Do \textbf{not} include next-step instructions in \texttt{"new\_content"}.

\textbf{Core Principles}:
\begin{itemize}[nosep, leftmargin=1.5em]
  \item Always preserve the original task description --- without it, the agent loses task context entirely.
  \item Preserve information the agent will need; remove or compress information it won't.
  \item If the agent is ignoring something important, make it more prominent.
  \item If the agent is repeating itself, ensure the context clearly shows the action was already taken and its outcome.
\end{itemize}

\textbf{General Techniques}:
\begin{itemize}[nosep, leftmargin=1.5em]
  \item Keep recent context detailed; compress older context progressively.
  \item Use clear structural markers to distinguish completed actions, findings, and constraints.
  \item Retain specific details likely to be needed later --- overly aggressive summarization can cause redundant work.
\end{itemize}

\smallskip\textbf{Your Input}

Agent Context: \texttt{\{\{full\_memory\}\}} \quad Token Usage Ratio: \texttt{\{\{token\_usage\_ratio\}\}}

\end{tcolorbox}
\end{table*}

\begin{table*}[ht]
\caption{\bcp\ judge prompt. \texttt{\{\{question\}\}}, \texttt{\{\{correct\_answer\}\}}, \texttt{\{\{actual\_answer\}\}}, and \texttt{\{\{explanation\}\}} are filled at runtime. The system turn is: \emph{``You are a precise evaluator. Always respond with valid JSON format as requested.''}}\label{tab:prompt_judge_bcp}
\begin{tcolorbox}[colback=blue!5, colframe=black]

You are an expert judge tasked with evaluating whether an agent's response to a question is correct.

\smallskip\textbf{Question}: \texttt{\{\{question\}\}}

\smallskip\textbf{Ground Truth Answer}: \texttt{\{\{correct\_answer\}\}}

\smallskip\textbf{Agent's Final Answer}: \texttt{\{\{actual\_answer\}\}}

\smallskip\textbf{Agent's Explanation}: \texttt{\{\{explanation\}\}}

\smallskip
Your task is to determine if the agent's output is correct based on the ground truth answer. Be strict and precise in your judgment.

\smallskip\textbf{Evaluation Criteria}:
\begin{enumerate}[nosep, leftmargin=1.5em]
  \item Extract the final answer primarily from the agent's final answer field.
  \item Use the explanation as supporting context that can clarify, refine, or contradict the final answer.
  \item Compare the agent's overall output with the ground truth answer.
  \item The agent's answer is correct \textbf{only if} it is semantically equivalent to the ground truth.
  \item Allow for minor variations in phrasing, but the core information must match exactly.
  \item For numerical answers, allow small rounding differences (within 1\% or 0.1 units).
  \item If the final answer contains additional information that does not contradict the ground truth, it can still be marked as correct.
  \item If the final answer is ambiguous, contradictory, or contains incorrect information, mark it as incorrect.
  \item If the agent did not provide a clear final answer, mark it as incorrect.
\end{enumerate}

\smallskip\textbf{Output Format}: Respond with a valid JSON object only (no additional text):

\begin{verbatim}
{
  "extracted_answer": "<exact answer extracted from the agent's output, or null>",
  "ground_truth":     "<the ground truth answer>",
  "reasoning":        "<why the answer is correct or incorrect>",
  "score":            1.0 or 0.0
}
\end{verbatim}

\end{tcolorbox}
\end{table*}

\clearpage
\clearpage
\onecolumn
\begin{center}
\captionof{table}{\mcpwiki\ judge prompt. \texttt{\{\{task\}\}}, \texttt{\{\{concrete\_task\_description\}\}}, \texttt{\{\{available\_tools\}\}}, \texttt{\{\{execution\_summary\}\}}, \texttt{\{\{final\_solution\}\}}, \texttt{\{\{total\_rounds\}\}}, and \texttt{\{\{dependency\_analysis\}\}} are filled at runtime. The system turn is: \emph{``You are an expert AI task execution evaluator. Score each dimension objectively based on evidence.''}}
\label{tab:prompt_judge_mcp}
\begin{tcolorbox}[breakable, colback=blue!5, colframe=black]
You are an impartial evaluator judging the quality of an AI agent's tool-based task execution.

\smallskip
You must assign scores \textbf{only based on evidence} from the task, solution, and tool usage. Your evaluation should be:
\begin{itemize}[nosep, leftmargin=1.5em]
  \item Objective (avoid being influenced by language fluency or formatting).
  \item Justified (include specific reasons tied to each score).
  \item Robust against bias (ignore narrative style, verbosity, or formatting polish).
\end{itemize}

\smallskip\textbf{Critical format rules}:
\begin{itemize}[nosep, leftmargin=1.5em]
  \item \textbf{Do not} penalize for output format (JSON, text, etc.) unless the task presented to agent explicitly requires it.
  \item If the task presented to agent says ``provide information'' without specifying format, \textbf{any} readable format is acceptable.
  \item Only deduct points for format if the task explicitly states ``return as JSON'' or ``format as table'' etc.
  \item Focus on \textbf{content} correctness, not presentation style.
\end{itemize}

\smallskip\textbf{Available tools} (\texttt{\{\{N\}\}} tools): \texttt{\{\{available\_tools\}\}}

\smallskip\textbf{Task presented to agent}: \texttt{\{\{task\}\}}

\smallskip\textbf{Concrete task reference} (for evaluation context only). Note: the agent did \textbf{not} see this concrete version, it only saw the task above. The task visible for the agent is the fuzzy version of the concrete task. This reference helps assess actual task completion but is not the sole criterion. The agent's interpretation of the fuzzy task may differ but still be valid.

\smallskip\textbf{Format reminder}: if the concrete task mentions JSON but the task presented to agent doesn't explicitly require it, \textbf{do not} penalize for not using JSON format. Only the task presented to agent's requirements matter for format. \texttt{\{\{concrete\_task\_description\}\}}

\smallskip\textbf{Execution summary}: \texttt{\{\{execution\_summary\}\}}

\smallskip\textbf{Final solution}: \texttt{\{\{final\_solution\}\}}.\quad\textbf{Total rounds}: \texttt{\{\{total\_rounds\}\}}.

\smallskip\textbf{Dependency analysis} (reference only). Note: this analysis was generated during task creation to help understand tool dependencies. The agent did \textbf{not} see this analysis. It is provided as reference for evaluation purposes. \texttt{\{\{dependency\_analysis\}\}}

\smallskip\textbf{Task completion rubric} (1--10 per subdimension).

\smallskip\textit{1. Task fulfillment and quality.}
\begin{itemize}[nosep, leftmargin=1.5em]
  \item 1--3: Perfectly completes 10--30\% of requirements.
  \item 4--6: Perfectly completes 40--60\% of requirements.
  \item 7--8: Perfectly completes 70--80\% of requirements.
  \item 9--10: Perfectly completes 90--100\% of requirements.
\end{itemize}
NOTE: requirements come from the task presented to agent only. Format (JSON/text) is \textbf{not} a requirement unless explicitly stated in the task presented to agent.

\smallskip\textit{2. Grounding.}
\begin{itemize}[nosep, leftmargin=1.5em]
  \item 1--3: 10--30\% of claims are perfectly grounded in tool outputs.
  \item 4--6: 40--60\% of claims are perfectly grounded in tool outputs.
  \item 7--8: 70--80\% of claims are perfectly grounded in tool outputs.
  \item 9--10: 90--100\% of claims are perfectly grounded in tool outputs.
\end{itemize}

\smallskip\textbf{Tool usage rubric} (1--10 per subdimension).

\smallskip\textit{1. Tool appropriateness.}
\begin{itemize}[nosep, leftmargin=1.5em]
  \item 1--3: 10--30\% of tools were perfectly selected for their subtasks.
  \item 4--6: 40--60\% of tools were perfectly selected for their subtasks.
  \item 7--8: 70--80\% of tools were perfectly selected for their subtasks.
  \item 9--10: 90--100\% of tools were perfectly selected for their subtasks.
\end{itemize}

\smallskip\textit{2. Parameter accuracy.}
\begin{itemize}[nosep, leftmargin=1.5em]
  \item 1--3: 10--30\% of tool calls have perfectly accurate and complete parameters.
  \item 4--6: 40--60\% of tool calls have perfectly accurate and complete parameters.
  \item 7--8: 70--80\% of tool calls have perfectly accurate and complete parameters.
  \item 9--10: 90--100\% of tool calls have perfectly accurate and complete parameters.
\end{itemize}

\smallskip\textbf{Planning effectiveness and efficiency} (1--10 per subdimension).

\smallskip\textit{1. Dependency awareness.}
\begin{itemize}[nosep, leftmargin=1.5em]
  \item 1--3: 10--30\% of dependency chains are perfectly executed.
  \item 4--6: 40--60\% of dependency chains are perfectly executed.
  \item 7--8: 70--80\% of dependency chains are perfectly executed.
  \item 9--10: 90--100\% of dependency chains are perfectly executed.
\end{itemize}

\smallskip\textit{2. Efficiency.}
\begin{itemize}[nosep, leftmargin=1.5em]
  \item 1--3: More than 70\% of tool calls are redundant or unnecessary.
  \item 4--6: 40--60\% of tool calls are redundant or unnecessary.
  \item 7--8: 10--30\% of tool calls are redundant or unnecessary.
  \item 9--10: Less than 10\% of tool calls are redundant or unnecessary.
\end{itemize}

\smallskip\textbf{Percentage-based scoring system}.

\smallskip\textbf{How to calculate scores}: for each dimension, calculate the defect rate $=$ (number of issues $/$ total opportunities) $\times$ 100\%. Then map defect rate to score:
\begin{itemize}[nosep, leftmargin=1.5em]
  \item 0--10\% defects $\to$ score 9--10 (excellent to perfect).
  \item 10--30\% defects $\to$ score 7--9 (good performance).
  \item 30--50\% defects $\to$ score 5--7 (average performance).
  \item 50--70\% defects $\to$ score 3--5 (poor performance).
  \item 70--100\% defects $\to$ score 0--3 (failed).
\end{itemize}

\smallskip\textbf{How to score}:
\begin{enumerate}[nosep, leftmargin=1.5em]
  \item When evaluating percentages, assess what counts as ``well executed'' for each dimension:
  \begin{itemize}[nosep, leftmargin=1.5em]
    \item Task fulfillment: requirements completed correctly.
    \item Grounding: claims supported by actual tool outputs.
    \item Tool appropriateness: suitable tools chosen for each subtask.
    \item Parameter accuracy: correct and complete parameters in tool calls.
    \item Dependency awareness: proper ordering of dependent operations.
    \item Efficiency: minimal redundant or unnecessary tool calls.
  \end{itemize}
  \item Minor imperfections reduce the percentage proportionally, not to zero.
  \item Map the resulting defect rate to the score range above.
\end{enumerate}

\smallskip\textbf{Key principles}:
\begin{enumerate}[nosep, leftmargin=1.5em]
  \item \textbf{Always} calculate as percentage, \textbf{not} absolute numbers.
  \item 10 errors in 100 calls (10\%) $=$ same score as 1 error in 10 calls (10\%).
  \item Consider the opportunity count for each dimension:
  \begin{itemize}[nosep, leftmargin=1.5em]
    \item Tool calls: how many total calls were made?
    \item Parallelization: how many tasks \textbf{could} have been parallel?
    \item Parameters: how many total parameters across all calls?
    \item Claims: how many factual statements were made?
    \item Dependencies: how many dependency relationships exist?
  \end{itemize}
\end{enumerate}

\smallskip
Score each dimension based on the defect rate mapped above. Use the full 1--10 range.

\smallskip\textbf{Concrete scoring examples with proportions}:
\begin{itemize}[nosep, leftmargin=1.5em]
  \item \textit{Task fulfillment}: 19/20 requirements (5\% defect) $=$ 9; 16/20 (20\%) $=$ 8; 12/20 (40\%) $=$ 6; 8/20 (60\%) $=$ 4.
  \item \textit{Tool appropriateness}: 19/20 optimal (5\% defect) $=$ 9; 16/20 (20\%) $=$ 8; 12/20 (40\%) $=$ 6; 8/20 (60\%) $=$ 4.
  \item \textit{Efficiency}: 2/20 redundant (10\%) $=$ 9; 4/20 (20\%) $=$ 8; 8/20 (40\%) $=$ 6; 12/20 (60\%) $=$ 4.
  \item \textit{Grounding}: 19/20 supported (5\% unsupported) $=$ 9; 16/20 (20\%) $=$ 8; 12/20 (40\%) $=$ 6; 8/20 (60\%) $=$ 4.
  \item \textit{Parameter accuracy}: 95/100 perfect (5\% defect) $=$ 9; 80/100 (20\%) $=$ 8; 60/100 (40\%) $=$ 6; 40/100 (60\%) $=$ 4.
  \item \textit{Dependency awareness}: 9/10 ordered (10\% misordered) $=$ 9; 8/10 (20\%) $=$ 8; 6/10 (40\%) $=$ 6; 4/10 (60\%) $=$ 4.
\end{itemize}

\smallskip\textbf{Format notes}:
\begin{itemize}[nosep, leftmargin=1.5em]
  \item Text output when JSON not required in the task presented to agent $=$ no penalty (0\% defect).
  \item Missing JSON when explicitly required in the task presented to agent $=$ count as failed requirement.
\end{itemize}

\smallskip\textbf{Normalize by complexity} (don't punish complex tasks):
\begin{itemize}[nosep, leftmargin=1.5em]
  \item Simple task: 1 error / 5 steps (20\% defect) $=$ score 7.
  \item Complex task: 4 errors / 20 steps (20\% defect) $=$ score 7.
  \item Server count is irrelevant --- using more servers is \textbf{not} better.
\end{itemize}

\smallskip\textbf{Critical evaluation requirements}:
\begin{enumerate}[nosep, leftmargin=1.5em]
  \item For task fulfillment, use chain-of-thought: first list \textbf{all} requirements from the task, then for each state whether fulfilled with evidence, then count fulfilled/total $=$ percentage, then map to score range.
  \item You \textbf{must} map each score to the exact percentage ranges in the rubrics.
  \item Task completion and tool usage \textbf{must} be evaluated against the concrete task reference, not the fuzzy task.
  \item Planning effectiveness should be evaluated based on the proportion of dependencies correctly handled, not the absolute number of steps executed or exact conformance to the dependency analysis.
  \item First calculate the actual percentage of completion/success, then assign the corresponding score range.
  \item Focus on completion \textbf{ratios} not absolute numbers --- completing 7/10 steps (70\%) should score similarly to completing 14/20 steps (70\%), regardless of task complexity.
\end{enumerate}

\smallskip
Please score based on completion percentages and proportional success, not absolute numbers. Return your evaluation scoring and reasoning in this exact JSON format. \textbf{All six numeric score fields (integer 1--10) are mandatory.} Missing any score field invalidates the response.

\begin{verbatim}
{
  "task_fulfillment_reasoning":
      "Explain how well the agent fulfilled the detailed task objectives,
       referencing specific content from the concrete task description
       and what percentage was completed.",
  "grounding_reasoning":
      "Explain how well the agent's outputs were grounded in actual
       tool results versus unsupported claims.",
  "tool_appropriateness_reasoning":
      "Explain whether the tools selected were appropriate for each
       subtask requirement.",
  "parameter_accuracy_reasoning":
      "Explain the accuracy and completeness of parameters used in tool
       calls, noting any missing required parameters or incorrect values.",
  "dependency_awareness_reasoning":
      "Explain how well the agent understood and respected task
       dependencies (what percentage of dependencies were handled
       correctly), refer to the provided dependency analysis section.",
  "parallelism_efficiency_reasoning":
      "Explain the efficiency of execution, including use of parallelism
       and avoiding redundancy, refer to the provided dependency analysis
       section.",
  "task_fulfillment":           X,
  "grounding":                  X,
  "tool_appropriateness":       X,
  "parameter_accuracy":         X,
  "dependency_awareness":       X,
  "parallelism_and_efficiency": X
}
\end{verbatim}

\smallskip
Return \textbf{only} the JSON object.
\end{tcolorbox}
\end{center}

\clearpage
\begin{table*}[ht]
\caption{Extraction prompt applied when \texttt{get\_document} (\bcp) or \texttt{get\_article} / \texttt{get\_sections} (\mcpwiki) results are appended to the context. \texttt{\{\{chunk\_idx\}\}}, \texttt{\{\{total\_chunks\}\}}, \texttt{\{\{question\}\}}, \texttt{\{\{existing\_notes\}\}}, and \texttt{\{\{chunk\}\}} are filled at runtime. If the tool result fits within the context budget it is processed in a single pass; otherwise it is split into overlapping chunks and the reading note is refined iteratively.}
\label{tab:prompt_extract}
\begin{tcolorbox}[colback=blue!5, colframe=black]

Objective: Your mission is to help answer the ``Original Question'' by refining a ``Reading Note'' based on sequentially provided document chunks. This iterative process ensures the note evolves into a concise and relevant tool for addressing the question.

\smallskip\textbf{Your Task}

Analyze the ``New Chunk of Document'' and critically revise the ``Existing Reading Note''. Your goal is to produce an updated version of the note, incorporating new insights while maintaining focus on the original question.

\smallskip\textbf{Key Guidelines}
\begin{enumerate}[nosep, leftmargin=1.5em]
  \item \textbf{Relevance is Key}: Include only the information that directly or potentially contributes to answering the ``Original Question''. Eliminate irrelevant or redundant details.
  \item \textbf{Refine, Don't Just Append}:
    \begin{itemize}[nosep, leftmargin=1.5em]
      \item \textbf{Merge}: Consolidate new information with existing points to enhance clarity and completeness.
      \item \textbf{Update}: Replace general statements with more precise or specific findings from the new chunk.
      \item \textbf{Remove}: Discard outdated, irrelevant, or less important sections of the note.
    \end{itemize}
  \item \textbf{No-Change Option}: If the new chunk provides no relevant information, simply return the unchanged Existing Reading Note.
  \item \textbf{Be Concise}: Keep the note succinct, capturing only the most critical and essential facts. Avoid summarizing the entire document; this is a working reference, not a comprehensive report.
  \item \textbf{No Premature Conclusions}: Focus strictly on refining the note at each step. Save final judgments or conclusions until all chunks have been processed.
\end{enumerate}

\smallskip\textbf{Output Format}

Your response must only consist of the full text of the revised ``Updated Reading Note''. Do not write any explanations, commentary, or other additional text.

\smallskip\rule{\linewidth}{0.4pt}

\textbf{Progress:} \texttt{\{\{chunk\_idx\}\}} out of \texttt{\{\{total\_chunks\}\}}

\smallskip\textbf{Original Question:} \texttt{\{\{question\}\}}

\smallskip\textbf{Existing Notes:} \texttt{\{\{existing\_notes\}\}}

\smallskip\textbf{New Chunk of Document:} \texttt{\{\{chunk\}\}}

\smallskip\textbf{Updated Reading Note:}

\end{tcolorbox}
\end{table*}

\end{document}